\pdfoutput=1

\documentclass[10pt, a4paper]{article}

\usepackage[final]{lrec2026} %

\usepackage[noautocite]{ai-usage-card}

\aiProjectName{\textcolor{colorA}{A}ffect, \textcolor{colorB}{B}ody, \textcolor{colorC}{C}ognition, \textcolor{colorD}{D}emographics, and \textcolor{colorE}{E}motion:\\ The \datasetName\ of Text Features for Computational Affective Science}
\aiDomain{Natural Language Processing}
\aiKeyApplication{Computational Affective Science}

\aiContactName{Jan Philip Wahle}
\aiContactEmail{wahle\@uni-goettingen.de}
\aiContactAffiliation{University of Göttingen}

\aiModels{GPT Codex 5.2, Claude Opus 4.5, Claude Opus 4}

\aiGeneratingCode{GPT Codex 5.2, Claude Opus 4.5, Claude Opus 4}
\aiRefactoringCode{GPT Codex 5.2, Claude Opus 4.5, Claude Opus 4}

\aiWhyUse{We used AI to produce source code for data processing.}
\aiMitigateErrors{We checked data quality and statistics separately and with human-in-the-loop.}
\aiMinimizeHarm{We manually verified the results the code produced.}

\name{Jan Philip Wahle$^*$$^\dagger$, Krishnapriya Vishnubhotla$^*$$^\ddagger$, Bela Gipp$^\dagger$, Saif M. Mohammad$^\ddagger$\\
  $^\dagger$University of Göttingen, Germany; $^\ddagger$National Research Council Canada\\
  \texttt{wahle@uni-goettingen.de; vkpriya@cs.toronto.edu, saif.mohammad@nrc-cnrc.gc.ca}\\ %
  $^*$equal contribution\\[1em]
  \hspace*{0.1cm}
  \begin{tabular}{l@{\hskip 0.5em}l@{\hskip 1em}l}
    \adjustbox{valign=c}{\includegraphics[height=1em]{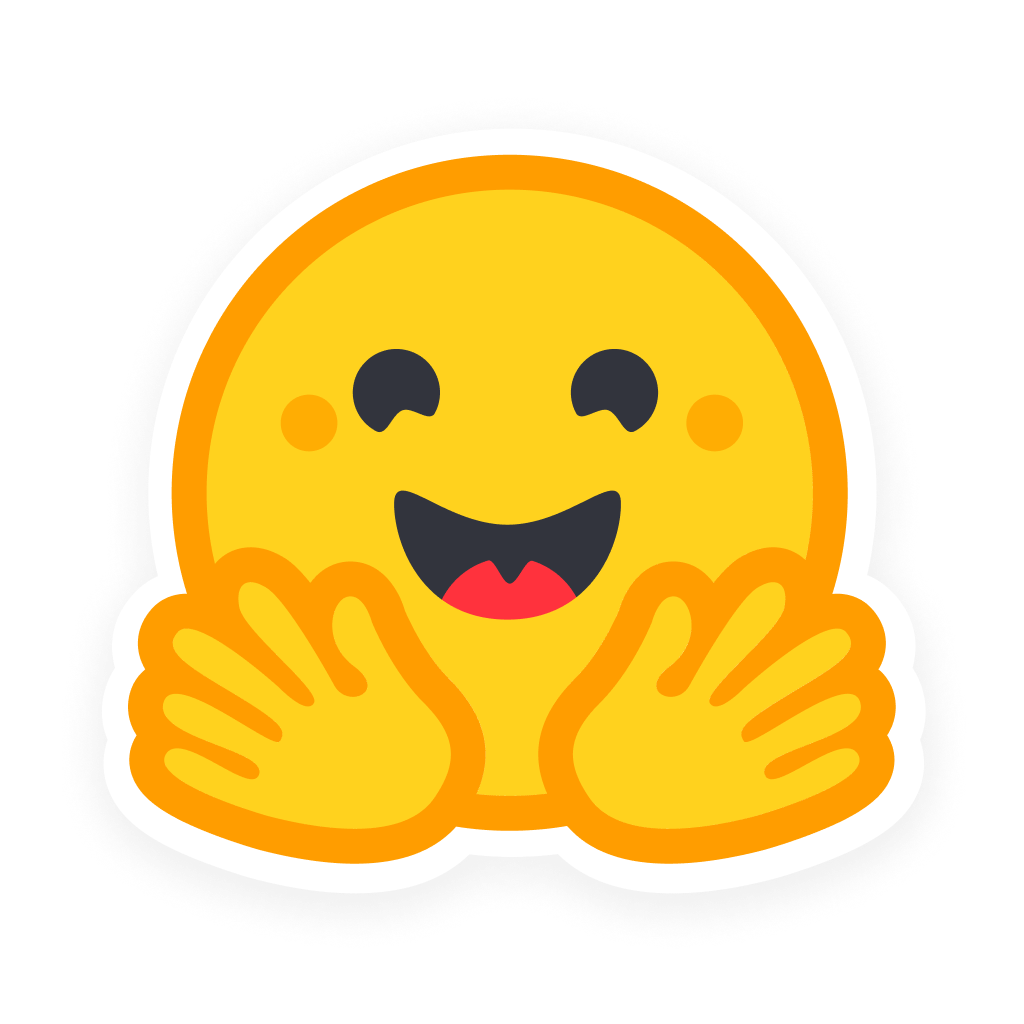}} & Dataset & \href{https://huggingface.co/datasets/jpwahle/abcde}{hf.co/datasets/jpwahle/abcde} \\
    \adjustbox{valign=c}{\includegraphics[height=1em]{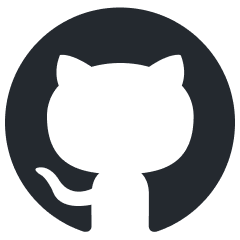}} & Code & \href{https://github.com/jpwahle/abcde}{github.com/jpwahle/abcde} \\
  \end{tabular}} 

\address{}

\usepackage[T1]{fontenc}

\usepackage[utf8]{inputenc}

\usepackage{microtype}

\usepackage{inconsolata}

\usepackage{graphicx}
\usepackage{booktabs}
\usepackage{cleveref}
\usepackage{longtable}
\usepackage{xspace}
\usepackage{xcolor}
\usepackage{array}
\usepackage{amssymb}
\usepackage{graphicx}
\usepackage{enumitem}
\usepackage{amssymb}
\usepackage{adjustbox} %
\usepackage{tabularx}
\usepackage{makecell}
\usepackage[noautocite]{annotation}

\definecolor{colorA}{HTML}{588BE0}
\definecolor{colorB}{HTML}{7083DA}
\definecolor{colorC}{HTML}{9775C1}
\definecolor{colorD}{HTML}{B46CA2}
\definecolor{colorE}{HTML}{C5658F}

\newcommand{\datasetName}{%
\textbf{%
\textcolor{colorA}{A}%
\textcolor{colorB}{B}%
\textcolor{colorC}{C}%
\textcolor{colorD}{D}%
\textcolor{colorE}{E}%
}\xspace
}

\title{\textcolor{colorA}{A}ffect, \textcolor{colorB}{B}ody, \textcolor{colorC}{C}ognition, \textcolor{colorD}{D}emographics, and \textcolor{colorE}{E}motion:\\ The \datasetName\ of Text Features for Computational Affective Science}

\abstract{
Work in Computational Affective Science and Computational Social Science explores a wide variety of research questions about people, emotions, behavior, and health. Often they make use of language data that is first labeled with relevant information such as the use of emotion words and age of the speaker. Even though many resources and algorithms exist to enable such labeling, finding and using them is still a substantial impediment, especially to practitioners in fields outside of computer science. 
Here, we present the \datasetName\ dataset (``\textcolor{colorA}{\textbf{A}}ffect, \textcolor{colorB}{\textbf{B}}ody, \textcolor{colorC}{\textbf{C}}ognition, \textcolor{colorD}{\textbf{D}}emographics, and \textcolor{colorE}{\textbf{E}}motion''), a large-scale collection of over 400 million released text instances from social media, blogs, books, and AI-generated sources, annotated for a number of features relevant to computational affective and social science. 
\datasetName facilitates inter-disciplinary research in wide range of fields, including affective science, cognitive science, the digital humanities, sociology, political science, and computational linguistics.
\\ \newline  }

\begin{document}

\maketitleabstract

\AddAnnotationRef{}

\section{Introduction}

Language is a rich medium of self-expression and communication.
It is a product of historical, cultural, embodied, and cognitive processes.  Consequently, computational approaches to the analysis of large text datasets have become a powerful tool for studying feelings, thought, body/health, and behavior of individuals and populations. %

\begin{figure}[t]
    \centering
    \includegraphics[width=.8\linewidth]{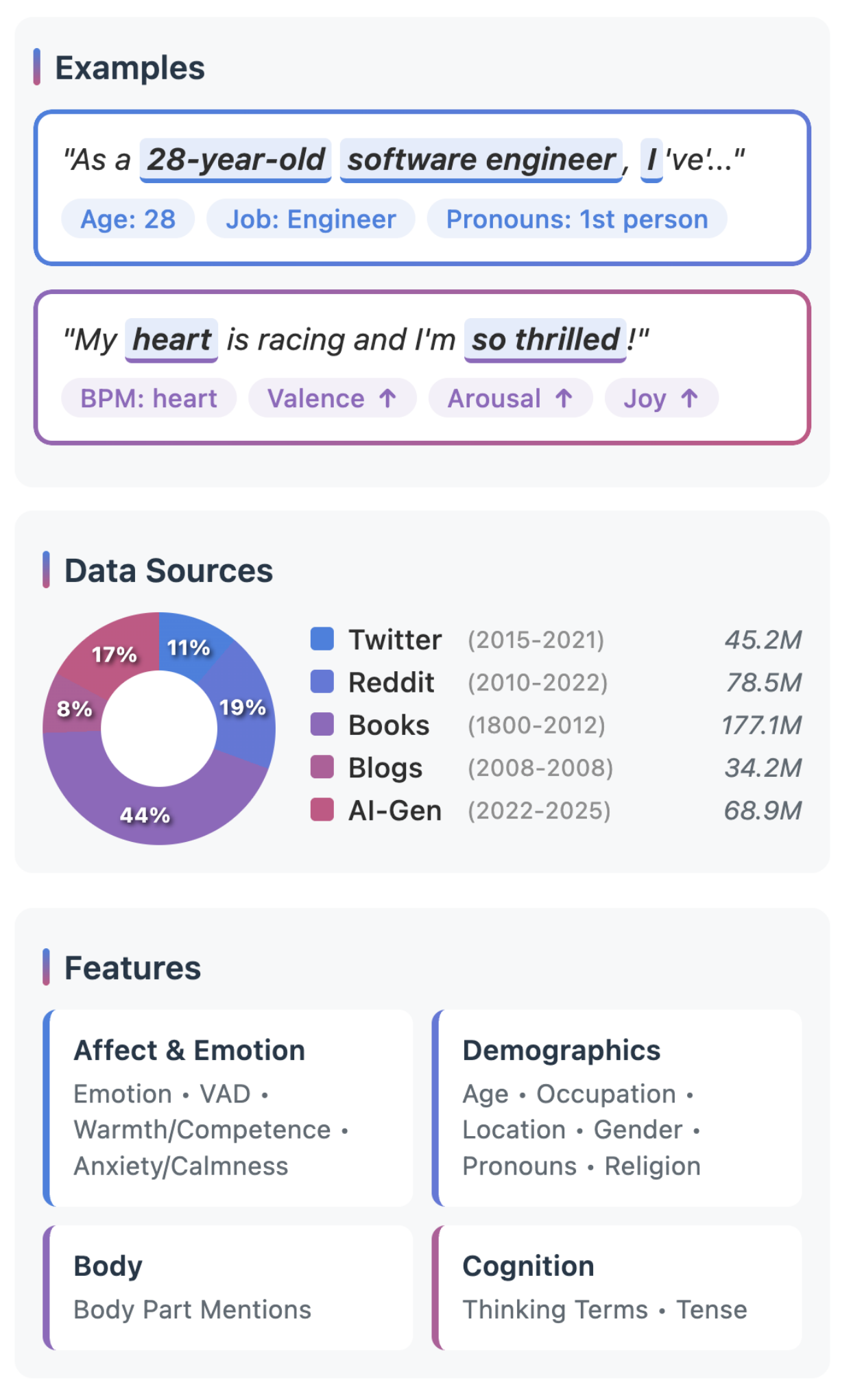}
    \caption{Overview of the \datasetName\ dataset.}
    \label{fig:teaser}
    \vspace*{-5mm}
\end{figure}

In the domain of Computational Affective Science (CAS), research has shown that language use is strongly associated with, and indicative of, mental states %
and cognitive processes in the human brain \cite{Guntuku2017DetectingDA}. When we are sad, we express ourselves using words %
associated with low valence (negative sentiment), or even subtle linguistic signals such as shorter, more terse utterances. 
When we lack confidence, we tend use more qualifiers such as \textit{``I think"}, \textit{``probably"}, and \textit{``right?"}. And so on. 
Linguistic properties of utterances are thus immensely useful signals of our mental models and processes, as well as how we as humans perceive ourselves and the world around us. 
At the same time, language is only an indirect trace of the underlying states we care about. Many mental states and cognitive processes are latent and must be inferred from behavior and context. Thus while it is difficult to determine the exact mental state of an individual from an utterance (something we do not recommend pursuing automatically anyway), the power of language analysis lies in bringing forth trends and patterns of mental states of whole populations (aggregate-level analysis). 
Language use remains a natural and scalable behavioral trace that %
can be used to
infer patterns of relative differences and change at an aggregate level. For example, how has the use of emotion words by a population changed with time and what psychological changes that indicates; how a cohort of people uses warmth and competence words when posting about different social groups; what moral frames are invoked when talking about controversial issues (such as immigration), etc.

Similarly, in the broader domain of social sciences, insights derived from textual data can complement those derived from qualitative methodologies like surveys and self-reports. Computational methods have the added advantage of enabling research using a much larger and more diverse pool of data when compared to lab studies with smaller groups of participants, thereby increasing the reliability of measurements. For example, Computational Social Science (CSS) researchers have used social media data to study mechanisms of influence and information diffusion in social networks \cite{goel2016structural}, monitor public sentiment and stance towards specific entities and topics \cite{dodds2011temporal}, identify changes in the contextualized semantic meanings of word types in different communities \cite{lucy2021characterizing}, etc.  

Another %
line of research that is of increasing interest to linguists, cognitive scientists, and computer scientists alike is the studying of AI-generated text. Machine learning based training of multi-billion parameter neural networks has resulted in Large Language Models that interface with and communicate in natural language with surprising fidelity \cite{achiam2023gpt}. Consequently, %
researchers are looking at the outputs of these models to understand statistical associations and biases that are encoded in them, particularly as LLMs take on a larger role in the data science pipeline as de-facto models for classification, labeling, and synthetic data generation \cite{ziems2024can, farrell2025large}. %
Researchers are also analyzing the "reasoning traces" or chains-of-thought (CoTs) of these models to better understand the artificial "cognitive" reasoning processes in these models, and how they compare with human processes \cite{korbak2025chain}.

All of this research that makes use of "language as data" relies on three key %
tasks as a pre-requisite to subsequent statistical analyses: collecting the textual data,
tagging it with associated metadata (extra-linguistic context), and the measurement of features of interest (quantifying, for example, sentiment from tweets). The dynamic nature of pipelines for data collection on the web, as well as the rapid pace of development of computational models of measurement for various constructs that are of interest in the social sciences, make this a technically challenging process for most researchers, especially those outside of computer science. Further, even for those comfortable with natural language processing, compiling and annotating a large number of diverse datasets and features is time and labor intensive.   

In this work, we compile and release \datasetName, a collection of text datasets automatically tagged with metadata associated with speaker demographics, and labeled for 136 lexical features that are of broad utility for researchers in the affective and social sciences. The selected datasets (aggregated from various primary sources) span multiple domains, genres, and time-periods. We group our text features into five thematic sets: \textcolor{colorA}{\textbf{A}}ffect, \textcolor{colorB}{\textbf{B}}ody, \textcolor{colorC}{\textbf{C}}ognition, \textcolor{colorD}{\textbf{D}}emographics, and \textcolor{colorE}{\textbf{E}}motion. 

Our primary motivation in creating this dataset is to unify textual features that encode aspects of behavior and cognition that are of interest to researchers in the affective sciences. While emotion and affect have been prominent topics of study here, recent advances in embodied cognition emphasize how bodily awareness and interactions with the environment shape many of these cognitive functions, and have downstream correlations with mental, emotional, and physical health \cite{wu2025language}. We therefore include body part mentions as a core feature of our dataset, allowing researchers to quantify patterns at the intersection of bodily mentions and emotions/affect. With Demographic information, we also consider features that encode \textit{social context}, in addition to linguistic and lexical features of the text itself. This is a crucial variable for most research in the affective and social sciences, where concepts like emotion and cognition are not constant, stable traits, but demonstrate significant variation with demographic features like age, within and across population sub-groups --- in fact, it is this variation that is often the phenomenon of interest \cite{hoemann2025construction,gutchess2023culture}.
Age-related differences have been reported in the arousal structure of emotion concepts \cite{trnka2022emotionconcepts}, and age and culture can jointly shift cognitive performance and style \cite{na2017ageculture}.

We make our resource publicly available for open access and use by researchers. We hope the \datasetName dataset will facilitate inter-disciplinary research in wide range of fields, including affective science, cognitive science, the digital humanities, sociology, political science, and NLP.

\section{Related Work}
The past several decades of research in computational linguistics and natural language processing has led to the development of tools that can be used to extract rich linguistic information from textual data, such as StanfordCoreNLP \cite{manning-etal-2014-stanford}, NLTK \cite{bird2009natural}, and spaCy\footnote{https://spacy.io/}. However, these are aimed largely at researchers with familiarity and expertise in these domains of research, with rich programming experience, and target features that are useful for linguistic analysis, like part-of-speech tags and dependency relations.
In the social sciences, by contrast, researchers are generally interested in quantifying more abstract, higher-level features of text, like sentiment or emotion intensity, political leaning, satire, literary quality, etc. \cite{licht2025measuring}   

There exists a small but growing set of accessible text analysis toolkits that have a narrower focus on specific domains of research. The Cornell Conversational Analysis Toolkit \cite{convo-kit} is a collection of datasets and computational scripts that enable research into aspects of social networks, conversational dynamics, and the sociolinguistics of online interactions. GutenTag is an NLP-driven tool for Digital Humanities research in particular, with a web-based interface that automatically extracts several features of interest from literary texts \cite{brooke-etal-2015-gutentag}.
In the Affective Sciences domain, packages like VADER \cite{Hutto2014VADERAP} and Textblob\footnote{https://textblob.readthedocs.io/} are widely used for sentiment analysis. The Emotion Dynamics toolkit \cite{vishnubhotla-mohammad-2022-tusc} provides scripts to quantify patterns of change in emotional expression in textual utterances over time.\footnote{\url{https://github.com/Priya22/EmotionDynamics}}

A prime example of a text processing toolkit that has seen broader use in the social, affective, and cognitive sciences, and the digital humanities, is LIWC (Linguistic Inquiry and Word Count), a proprietary software that quantifies several psychometric properties of words to facilitate studying the links between language, cognition, and psychology \cite{boyd2022development}.   

We position \datasetName as not just a feature extraction tool, but a much more comprehensive repository of data mapped to metadata and pre-computed features. We compile, clean, and annotate multiple text datasets from diverse domains that are of broad interest to affective and cognitive science researchers, including social media, blogs, and AI-generated text. By pre-computing features for these datasets, our resource lowers the technical barrier of entry for many researchers, and also encourages reproducibility of empirical research in the field.
The \datasetName dataset therefore functions as a much more thorough and standardized starting point, on top of which researchers can directly apply statistical methods of analysis in order to answer various research questions.

\section{The \datasetName\ Dataset}

We compiled a large-scale, longitudinal collection of textual datasets from various primary sources, spanning data from social media platforms, books, and blogs. Additionally, we compiled a selection of AI-generated text, including human--LLM conversational data, LLM reasoning traces, preference datasets, and datasets from AI-generated text detection tasks. We then computed and recorded the text features enumerated in Section \ref{feature-desc} for every text \textit{instance} in each of these datasets.  Note that a text instance can be defined at different levels of granularity --  features can be recorded at the sentence-level (with each sentence in turn linked to the original Reddit post, blog post, AI-generated story, etc.), or for equal-sized chunks of tokens from each dataset, among others. Our definition of what constitutes a text instance differs from one dataset to another, and is enumerated in Section \ref{data-sources} alongside the dataset descriptions. In Section \ref{feature-desc}, we expand on our annotated features and the methods used to quantify them given a text instance. 

\subsection{Data Sources}
\label{data-sources}
We annotated datasets spanning millions of records from Twitter, Reddit, blogs, books, and AI-generated content:

\noindent\textbf{Twitter (TUSC, 2015--2021)}: 45.2M geolocated tweets sourced from the TUSC dataset \cite{vishnubhotla-mohammad-2022-tusc}. Each tweet is considered an instance.

\noindent\textbf{Reddit (Pushshift, 2010--2022)}: 78.6M posts crawled from Reddit via Pushshift archives hosted by the Internet Archive \cite{baumgartner2020pushshift}. We remove posts that are marked as adult (over\_18), promoted, containing images or videos, and having fewer than 5 or more than 1,000 words; each post constitutes a text instance.   

\noindent\textbf{Books (Google, 1800--2012)}: 177.1M 5-gram occurrences (1.74M unique 5-grams) from the English Fiction subset of the Google Books Ngram Corpus (version 20120701) \cite{googlebooks2012}. Each 5-gram is treated as an instance. While these instances are short, they remain useful for large-scale frequency and longitudinal analyses.

\noindent\textbf{Blogs (Spinn3r, 2008)}: 34.2M personal blog entries from the ICWSM 2009 conference \cite{burton2009icwsm}. Each blog post is an instance.

\noindent\textbf{AI-Generated Texts (Various, 2022--2025)}: 68.9M AI-completions from 15 distinct datasets, including conversational texts (WildChat-1M \cite{zhao2024wildchat}, LMSYS-Chat-1M \cite{zheng2023lmsys}, PIPPA \cite{pippa2024}, HH-RLHF \cite{hh-rlhf2024}, Prism \cite{prism2024}, and APT \cite{wahle-etal-2022-large}; persuasive essays (Anthropic-Persuasiveness \cite{anthropic2024persuasion}); AI text detection datasets (M4 \cite{wang-etal-2024-m4}, MAGE \cite{li-etal-2024-mage}, LUAR \cite{sotofew}); reasoning traces (General Thoughts 430k \cite{generalthoughts2024}, Reasoning Shield \cite{reasoningshield2024}, SafeChain \cite{jiang2025safechainsafetylanguagemodels}, STAR-1 \cite{wang2025star}); and narratives (TinyStories \cite{eldan2023tinystories}). For conversational datasets, we define the LLM response from a single user--chatbot interaction turn as a text instance (multi-turn conversations are split into multiple instances). Preference datasets generally comprise of a pair of LLM generations (in response to a user prompt) along with an indicator of the preferred generation; we take each generation separately to be an instance of AI-generated text. The AI-generated text detection datasets consist of LLM generations of varying lengths, intended to simulate human text in specific domains like news articles and social media posts; each generation is considered an instance. For reasoning traces, we disregard the final model output and only use the chain-of-thought text for each query as the instance. Instances are multi-feature. One text can activate several feature families, and some short texts may not activate any feature in a given family.

\subsection{Annotated Features}
\label{feature-desc}

We extract linguistic features through a set of lexicons, word lists, and regular expressions, organized along five dimensions relevant to computational affective science: Affect, Emotion, Body, Cognition, and Demographics. Word lists and lexicons as a measurement method offer several advantages, particularly in inter-disciplinary studies: ease-of-use, generalization power, reliability, adaptability, and interpretability, among others. Variation in word choices across populations and data subsets is also intertwined with multiple cognitive and social processes relating to language use (such as power dynamics, cultural and demographic factors, medium of communication, etc.). Appendix~\ref{app:feature-summary} summarizes the released resources, and Appendix~\ref{app:dataset-stats} reports per-source text statistics.

Affect and Emotion features are characterized using word association lexicons; Body and Cognition features through a curated list of terms, compiled from multiple sources, that indicate a strong association with body and cognition-related activities; and Demographic features through hand-constructed regular expressions and heuristic rules. For the demographic feature "Occupation", for example, we use a regex to parse self-disclosure statements of the type "I am/work at/employed as \textit{[occupation]}." from a user, and apply the extracted term to all text instances from that user. The demographic feature "Age", which refers to the age of the user at the time of posting, is computed by first using a similar regex for age or date-of-birth, and then applying a heuristic rule that combines this with the timestamp of the post. 

For each feature dimension (say, valence), and each text instance (say, a tweet), we use the associated word-level lexicon to compute an aggregate instance-level score in multiple ways: the \textbf{average} valence score of the constituent terms, a binary \textbf{flag} indicating if a word from the valence lexicon is present in the instance, and the \textbf{count} of the number of lexicon terms present in the instance.
The appropriateness of a particular aggregate measure for a feature will depend on the specifics of the research question being answered, and we leave this decision to the users. For example, length-normalized count features are more appropriate as a comparative indicator for texts of different lengths, rather than the average intensity score.

\noindent {\bf Affect:} Affect refers to the fundamental neural processes that broadly determine and regulate internal experiences of emotion, mood, and feelings. Affective states are generally characterized along three principal dimensions: valence (scale of positive--negative), arousal (scale of active--passive), and dominance (scale of competent--incompetent / powerful--weak), together referred to as VAD, which form our feature dimensions for Affect.
We match words against the NRC VAD lexicon \cite{vad-acl2018}, which maps words to a real-valued intensity score between 0 (lowest) and 1 (highest). We further define `High' and `Low' VAD features, where only lexicon entries with scores $\geq0.66$ and $\leq0.33$ respectively are considered.

\noindent {\bf Emotion:} We compute average, count, and binary presence flag features for multiple %
discrete emotions: the eight basic emotions of anger, anticipation, disgust, fear, joy, sadness, surprise, and trust, computed using the NRC Emotion Intensity lexicon \cite{LREC18-AIL}; warmth, competence, sociability, and trust using the WCST Lexicon \cite{mohammad2025words}; and anxiety and calmness with the NRC WorryWords lexicon \cite{worrywords-emnlp2024}.%

\noindent {\bf Body}: We identify Body Part Mentions (BPMs) by matching text tokens against curated lists of 292 anatomical unigrams, bigrams, and trigrams from \citet{zhuang-etal-2024-heart, wu2025language}, capturing mentions with possessive pronouns (e.g., ``my heart'', ``her hand''). Features are recorded as binary flags for the presence of a body part mention, and for each possessive pronoun, as a list of BPMs used with that pronoun (i.e, \texttt{MyBPM} is a list containing BPMs used with the pronoun `my'). Appendix~\ref{app:bpm} lists the body part words used.

\begin{figure*}[t!]
    \centering
    \includegraphics[width=.88\linewidth]{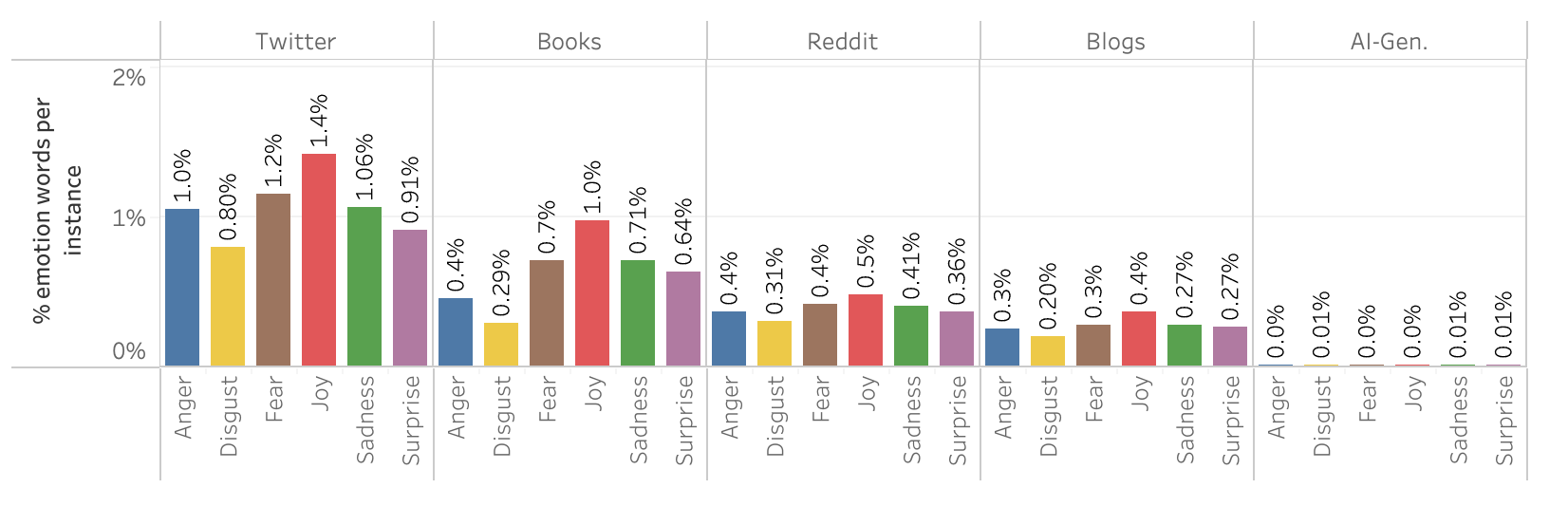}
    \caption{\textbf{Q1. Emotion:} Percent of words per instance from the six Ekman emotion categories.} %
    \label{fig:ae-emotion}
\end{figure*}

\begin{figure}[t!]
    \centering
    \includegraphics[width=\linewidth]{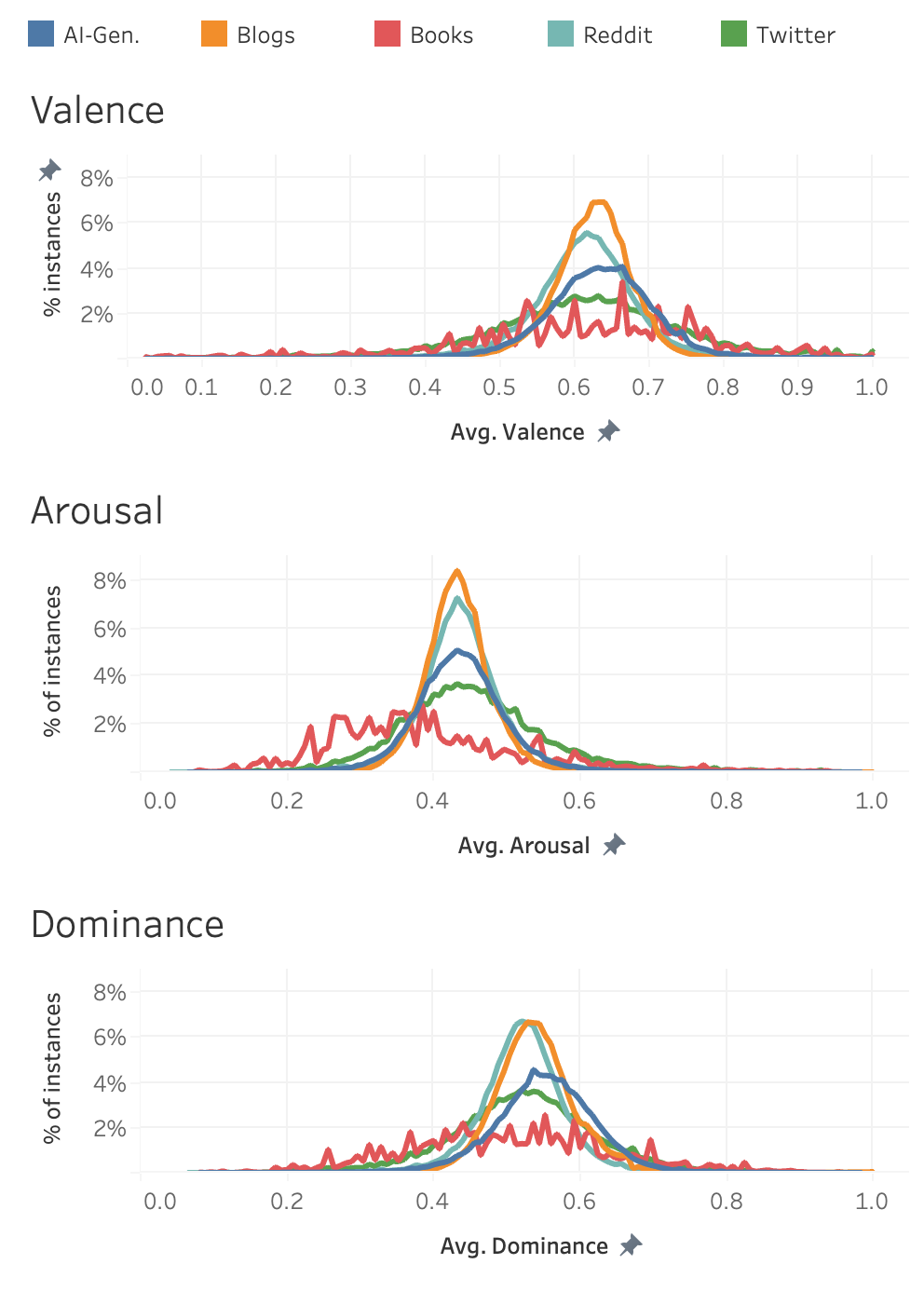}
    \caption{\textbf{Q1. Affect:} Distributions of average Valence, Arousal, and Dominance (VAD) scores of all words per instance, by source.}
    \label{fig:ae-vad}
\end{figure}

\noindent {\bf Cognition}: We classify cognitive processes by identifying ``thinking words'' derived from categories outlined in Bloom’s Taxonomy and Queensland’s Glossary of Cognitive Verbs.\footnote{\url{https://adp.uni.edu/documents/bloomverbscognitiveaffectivepsychomotor.pdf}}$^{,}$\footnote{https://www.qcaa.qld.edu.au/downloads/senior-qce/common/snr\_glossary\_cognitive\_verbs.pdf} 
We categorize 98 unigrams into 11 categories (e.g., analyzing, learning, decision-making). A binary flag feature indicating presence is recorded for each category. Appendix~\ref{app:cog} lists the full grouped inventory.

\noindent {\bf Demographics:} We extract demographic attributes, including age, occupation, gender, country, city, and religion, using regular expression matching and structured dictionary lookups. We map occupations according to the U.S. Bureau of Labor Statistics Standard Occupational Classification (SOC) system, and match gender, country, city, and religion references against controlled vocabularies from Wikipedia and Geonames datasets.\footnote{ https://en.wikipedia.org/wiki/List\_of\_gender\_identities, List\_of\_religions\_and\_spiritual\_traditions,List\_of\_countries\\\_and\_dependencies\_by\_population,www.bls.gov/soc/2018, https://public.opendatasoft.com/explore/dataset/geonames-all-cities-with-a-population-1000/}
These regexes are not exhaustive but target high-precision first-person self-disclosure patterns and public controlled vocabularies.
Appendix~\ref{app:dmg-regex} shows the full list of patterns used.
\\
\noindent\textbf{Focus features:} We also identify pronouns (possessive and non-possessive) and verb tenses (past, present, future) through direct lexical matching and morphological tagging using the English UniMorph dataset.\footnote{\url{https://github.com/unimorph/eng}} This enables quantifying linguistic usage patterns related to personal references and temporal framing.

\subsection{Lexical vs. ML Features} 

The measurement of a construct such as "anger" or "cognitive activity" given a text can be operationalized as a feature in many ways, and is dictated by the requirements of downstream use-case(s). If the goal is to predict the level of anger expressed in a single utterance by a person in real-time, trained machine learning models or pre-trained large language models will be more accurate than word-level lexicon aggregates. If, on the other hand, our goal is to measure the changes in anger levels expressed on Twitter in the USA over the last decade, plotting the (normalized) density of usage of anger-associated words in tweets from each year is a valid measurement method.

In other words, while lexicon-based methods are not as accurate as neural models at instance-level estimation (i.e, for estimating the sentiment of a particular sentence), they are comparable and sufficient when used as tools of aggregate-level analysis. This means that relative patterns of change, for comparing emotional intensity across different sources, or quantifying the patterns of changes in sentiment over a temporal period (longitudinal analysis), can be estimated with high accuracy using lexicon-based methods.
The source resources also provide validation evidence for aggregate use \cite{vad-acl2018,Mohammad13,zhuang-etal-2024-heart}.

In \citet{teodorescu2023evaluating}, the authors empirically demonstrate that emotion arcs (i.e, temporal fluctuations of valence intensity) with word-level lexicons of valence highly correlate with the ground-truth arcs (with correlation scores $>0.9$) for social media texts, provided certain hyper-parameters (like the width of the temporal window) are appropriately set. In the domain of digital humanities, \citet{ohman2024emotionarcs} measure the agreement of lexicon-generated arcs for classic fiction novels with human annotation, and find high consensus. In their work, the NRC emotion intensity lexicon was customized to the literary domain with a word-similarity measure, which is one of the recommended practices for the use of emotion lexicons \cite{Mohammad23ethicslex}.

\section{Key Research Questions}
We use  \datasetName\ to shed light on nine research questions, grouped by feature family.\\[2em]
\noindent\textbf{Q1. Affect--Emotion (AE):} \textit{%
To what extent are affect- and emotion-associated words %
used in different %
media?}\\ %
\textit{Motivation.} Affect and emotion lexicons provide aggregate signals about how people express emotion and affect across domains. These are central to population-level work in affective and social science to trace well-being, emotional contagion, or cultural mood shifts. For NLP, these signals offer interpretable, domain-portable emotion features for sentiment modeling and model alignment.\\
\textit{Results.} \Cref{fig:ae-emotion} shows the length-normalized counts (i.e, instance-level word density) of lexicon terms for six %
core emotion categories. Observe that Twitter has the highest %
emotion word usage rates (e.g., $\sim$1.4\% joy tokens per instance), followed by Books and Reddit. AI-generated text uses explicit emotion words rarely ($\approx$0.01\%). \Cref{fig:ae-vad} shows the distributions of average intensity scores for Valence, Arousal, and Dominance (VAD): human-authored sources center around neutral-to-positive valence (roughly 0.60--0.65), moderate dominance, and mid-to-low arousal. Books and AI-generated text display slightly lower arousal. %

\begin{figure}[t!]
    \centering
    \includegraphics[width=0.6\linewidth]{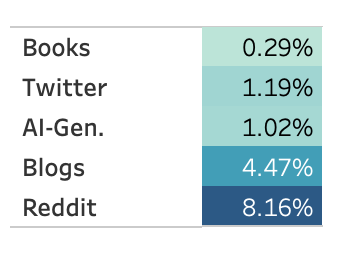}
    \caption{\textbf{Q2. Body:} Percent of instances with a positive flag for possessive body part mentions (e.g., \textit{my head/heart/}etc.) by source.}
    \label{fig:b-overall-dist}
\end{figure}

\begin{figure*}[t!]
    \centering
    \includegraphics[width=\linewidth]{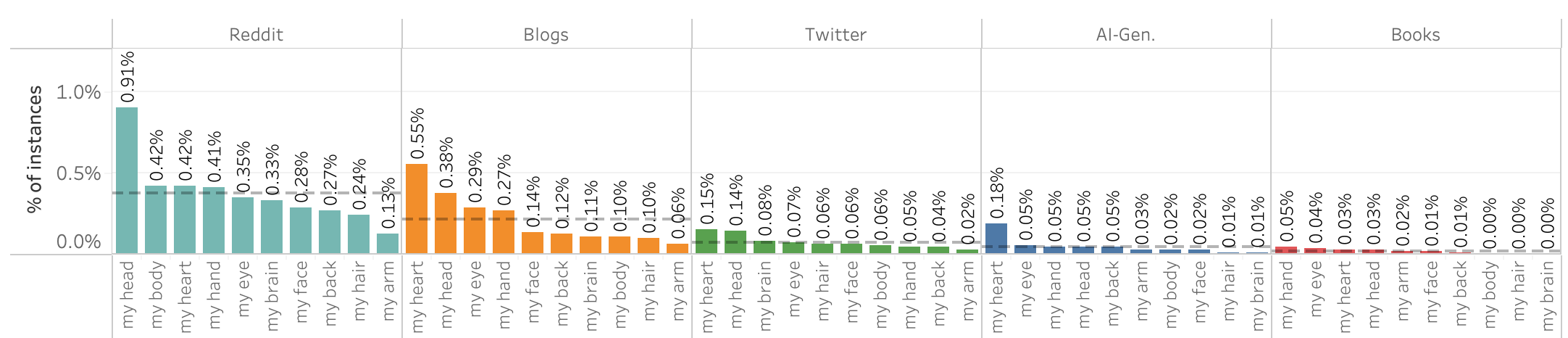}
    \caption{\textbf{Q2. Body:} \% of possessive BPMs for the top ten most common BPMs, across sources. Dashed lines represent the average occurence of BPMs per dataset.}
    \label{fig:b-individual-dist}
\end{figure*}

\begin{figure*}[t!]
    \centering
    \includegraphics[width=\linewidth]{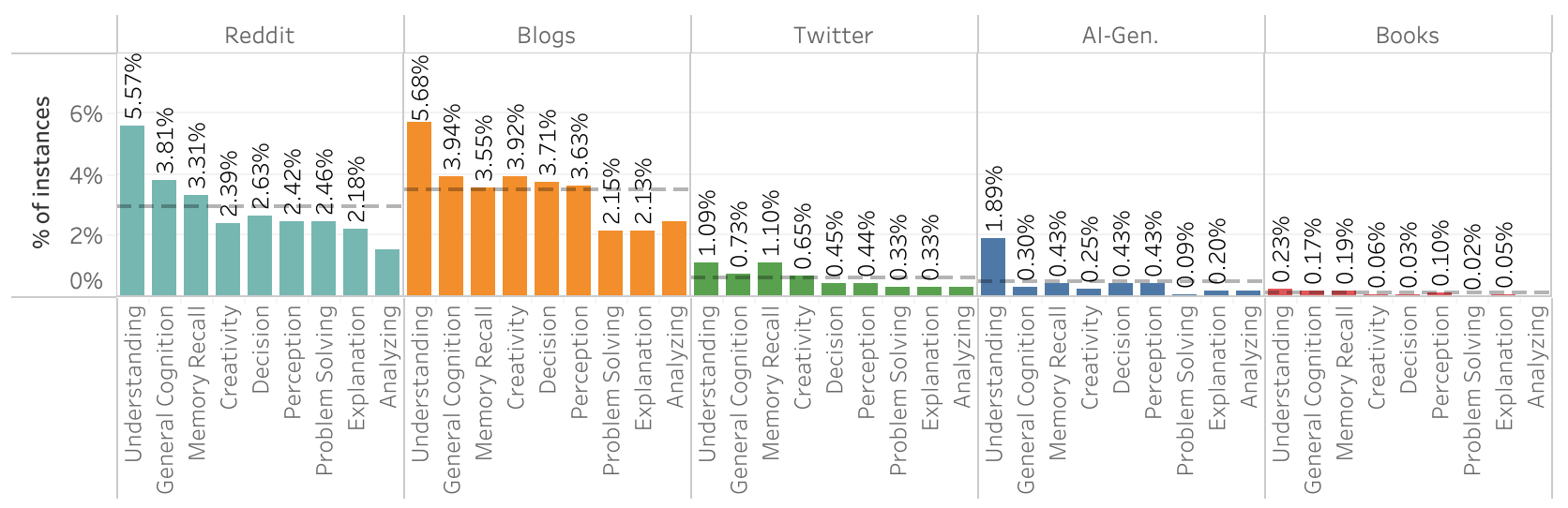}
    \caption{\textbf{Q3. Cognition:} \% of instances with a positive flag for cognitive terms from different categories. Dashed lines represent the average occurence of cognitive terms per dataset.}
    \label{fig:c-cog-word-dist}
\end{figure*}

\begin{figure}[t!]
    \centering
    \includegraphics[width=0.85\linewidth]{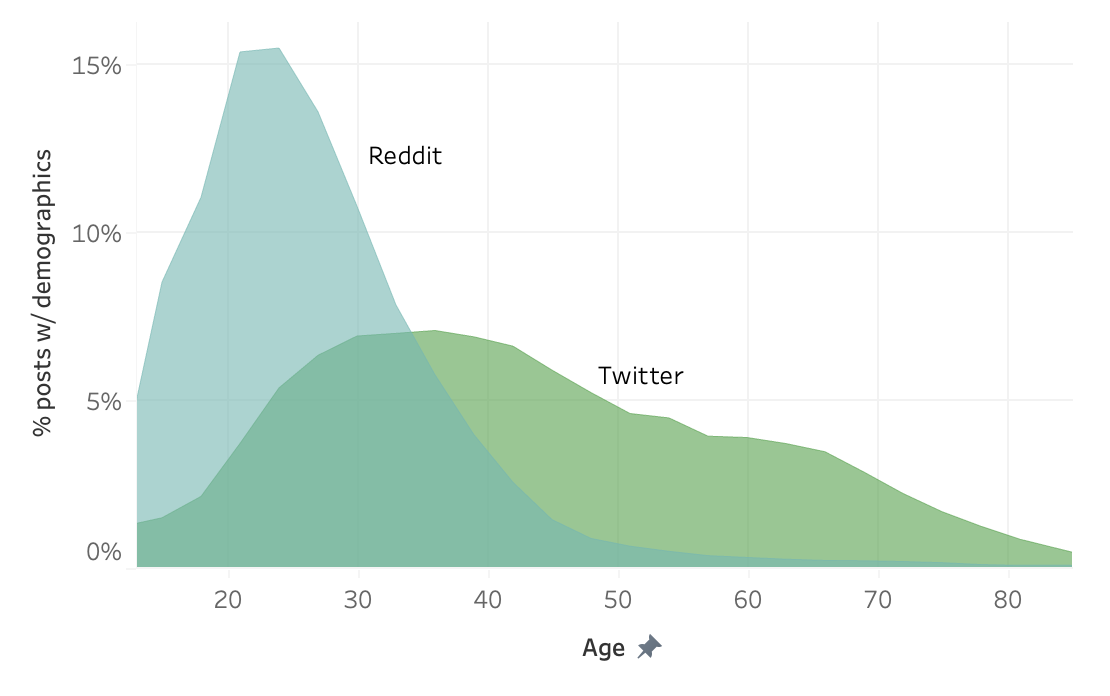}
    \caption{\textbf{Q4. Demographics:} Age distribution of posts with self‑disclosed age.}
    \label{fig:d-age}
\end{figure}

\noindent\textbf{Q2. Body (B):} \textit{How often do people refer to body parts in text? Does it vary in different media?}\\
\textit{Motivation.} Body parts mentioned (BPMs) can reveal how people linguistically connect to their bodies and thus how embodied experience manifests in text. In CSS, BPMs allow measurement of health, stress, or embodied metaphors in everyday discourse. In NLP, they can serve as features for affective and health-related modeling.\\
\textit{Results.} \Cref{fig:b-overall-dist} shows that Reddit (8.16\%) and Blogs (4.47\%) contain the most possessive BPMs (references to `\textit{my <BPM>}'); Books and AI-generated text contain very few. \Cref{fig:b-individual-dist} highlights the most frequent first-person BPMs: \textit{my head} is most common (especially on Reddit with 0.91\%), while \textit{my heart} appears more in Blogs (0.55\%). Mentions like \textit{my body, my face,} and my \textit{hands} vary systematically with domain. %

\noindent\textbf{Q3. Cognition (D):} \textit{How prevalent are cognition (``thinking'') words across domains, and which sub-processes are most/least common?} %

\noindent \textit{Motivation.} Cognition vocabulary (understanding, remembering, deciding, etc.) offers a linguistic window into reasoning and mental-state discourse. For CSS, this enables cross-platform comparison of analytical concepts (and one can draw relations to emotional language). For NLP, such features open a window into the interpretability of models of deliberation, argumentation, and education.\\
\textit{Results.} \Cref{fig:c-cog-word-dist} shows that human-authored sources use these terms far more than AI-generated text or Books. Blogs and Reddit lead overall: words of \emph{understanding} (5.68\%/5.57\%), \emph{general cognition} (3.94\%/3.81\%), and \emph{memory recall} (3.55\%/3.31\%). Twitter shows the same ranking  of categories, but at lower overall rates (e.g., understanding $\sim$1.09\%). AI-generated text underexpresses cognition terms across categories (e.g., problem solving $\sim$0.09\%, explanation $\sim$0.20\%). Across corpora, %
\emph{analyzing} is the %
least frequent cognition term in our set. %

\noindent\textbf{Q4. Demographics-Age (D-Age):} \textit{What are the self-disclosed age distributions in different media?} %

\noindent \textit{Motivation.} Age is a key demographic in CSS for understanding generational differences in discourse, politics, or emotion, and in NLP for bias control and stratified sampling. Self-disclosed age %
data helps characterize who participates in online discourse and calibrate downstream demographic analyses.\\
\textit{Results.} \Cref{fig:d-age} shows that 
late teens to late 20s are most common in Reddit, whereas 
the 30s are most common in Twitter. Tails decline steadily thereafter. %
This suggests that Reddit users are %
markedly younger than Twitter users. %

\begin{figure*}[t!]
    \centering
    \includegraphics[width=\linewidth]{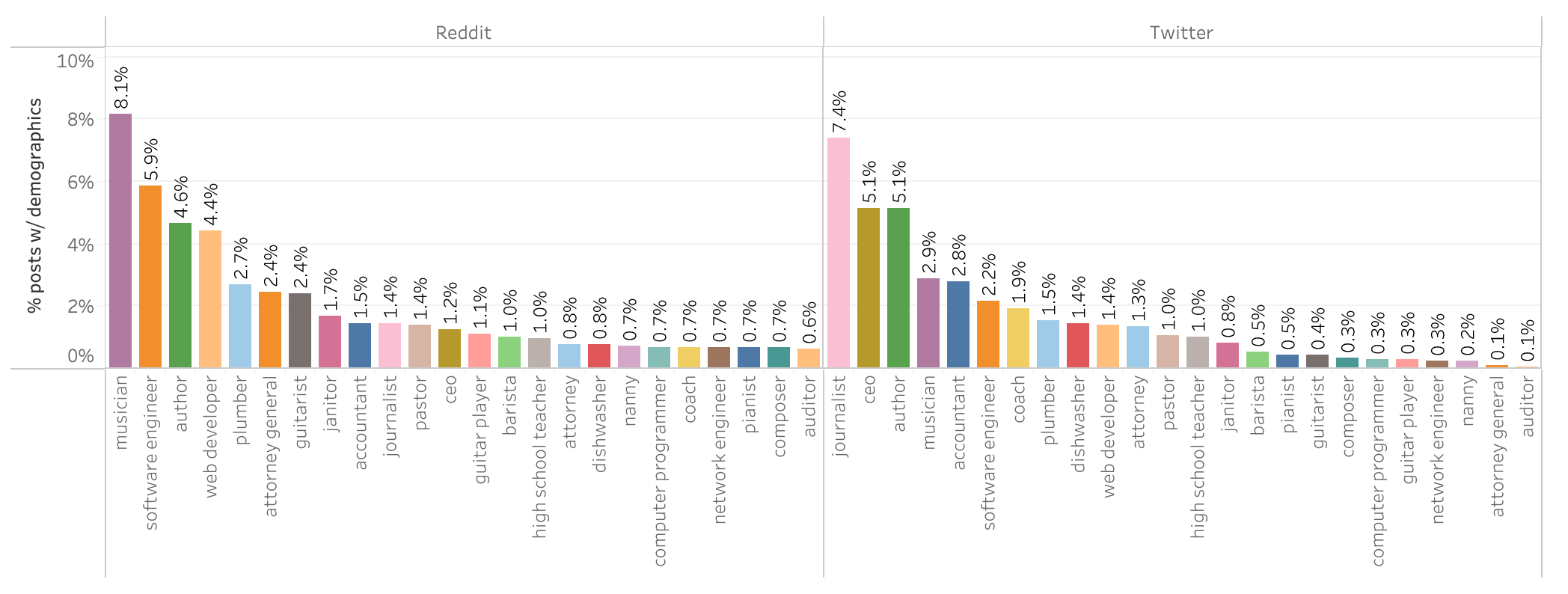}
    \caption{\textbf{Q5. D:} Occupation distribution of posts with self-disclosed occupation.}
    \label{fig:d-occupation}
\end{figure*}

\begin{figure}[t!]
    \centering
    \includegraphics[width=\linewidth]{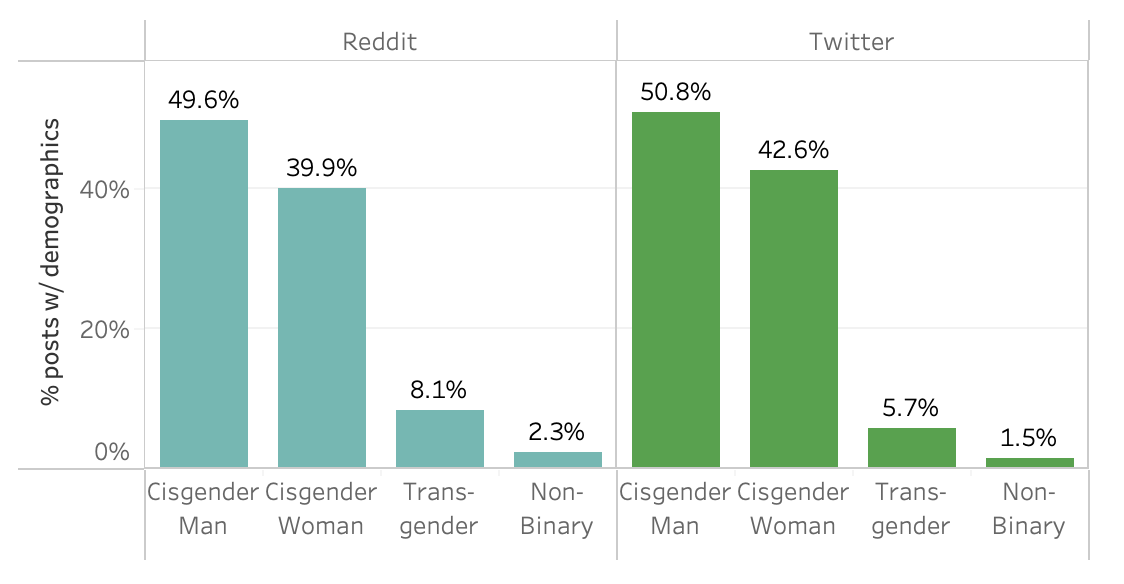}
    \caption{\textbf{Q6. D:} Gender distribution of posts with self-disclosed gender.}
    \label{fig:d-gender}
\end{figure}

\begin{figure}[t!]
    \centering
    \includegraphics[width=\linewidth]{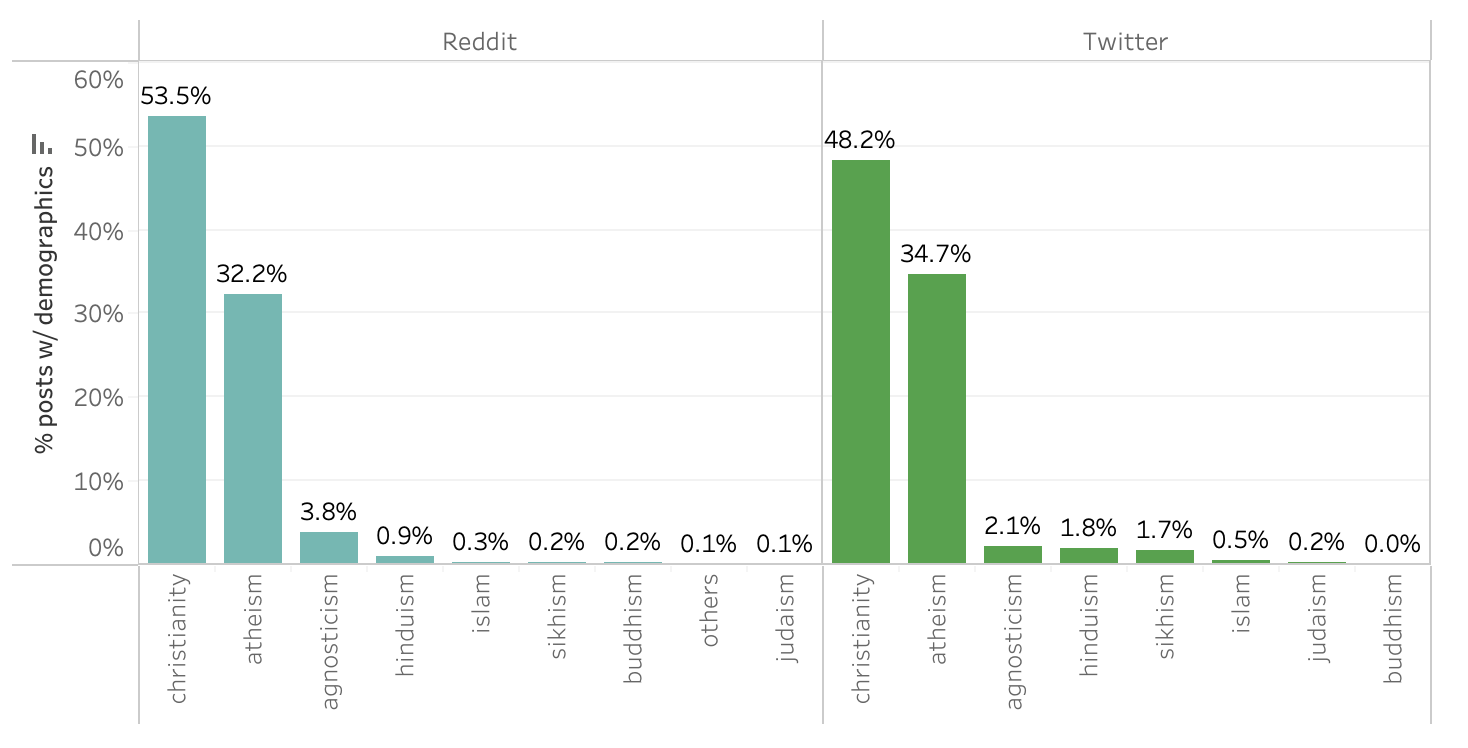}
    \caption{\textbf{Q7. D:} Religion distribution for posts with self-disclosed faith.}
    \label{fig:d-religion}
\end{figure}

\noindent\textbf{Q5. D-Occupation:} \textit{Which occupations are most frequently self-disclosed?}\\
\textit{Motivation.} In most modern societies, occupations are key parts of people's identity, values, and social interaction. In CSS, they enable studies of professional discourse, social stratification, and belief-linked behavior. In NLP, such self-labels offer interpretable subgroup features for fairness and personalization.\\
\textit{Results.} \Cref{fig:d-occupation} shows Reddit users often self-identify as creative or technical professionals (e.g., musicians, software engineers), whereas Twitter skews toward public-facing roles (e.g., journalists, authors, executives).

\noindent\textbf{Q6. D-Gender:} \textit{Which genders do users most often self-disclose?}\\
\textit{Motivation.} Gender is a central dimension of social identity (and inequality). Self-identified gender in text provides a precise, ethically grounded way to study participation patterns, self-representation, and language variation without resorting to gender inference or heuristics. In CSS, this enables analysis of representation and discourse norms; in NLP, it supports fairness auditing in generation, pronoun resolution, and toxicity detection.\\
\textit{Results.} \Cref{fig:d-gender} shows the majority of people self-disclose as cis men and women, with smaller but notable transgender (5--8\%) and non-binary (1--2\%) self-descriptions. These values highlight the presence of gender-diverse voices in online spaces and illustrate the potential for intersectional analyses when combined with other attributes (e.g., affect, occupation). Importantly, these reflect rates among disclosed posts, not population estimates. For downstream use, such disclosures serve as positive-only labels—absence of gender mention should be treated as missing, not negative.%

\noindent\textbf{Q7. D-Religion:} \textit{With which religions do users most often identify?}\\
\textit{Motivation.} Religious self-identification is an important dimension of cultural identity and social behavior. In CSS, religion enables studies of moral language, group cohesion, and polarization; in NLP, it is critical for evaluating fairness, bias, %
or sentiment models concerning faith-related content.\\
\textit{Results.} \Cref{fig:d-religion} shows that users mostly identify with Christianity and atheism on both Reddit and Twitter. On Reddit, Christianity accounts for 53.5\% of religion-disclosing posts and atheism for 32.2\%; on Twitter, Christianity (48.2\%) and atheism (34.7\%) again lead. Agnosticism follows (Reddit 3.8\%, Twitter 2.1\%), with smaller communities representing Hinduism, Sikhism, Islam, Judaism, and Buddhism (each under 2\%). Twitter exhibits relatively greater religious diversity (notably Hinduism and Sikhism) than Reddit. %

\begin{figure}[t!]
    \centering
    \includegraphics[width=\linewidth]{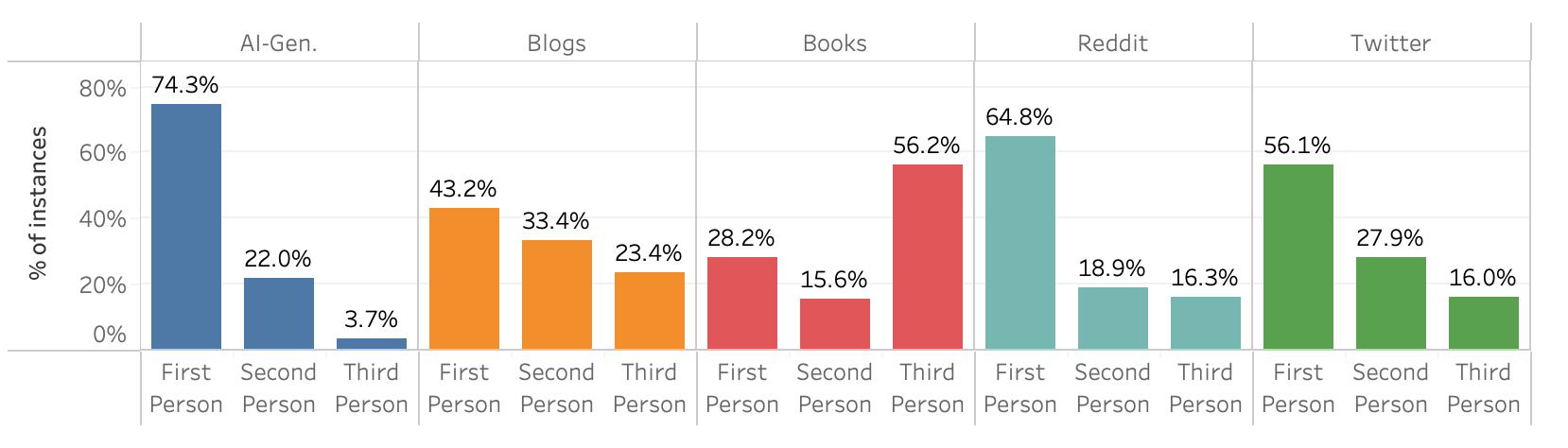}
    \caption{\textbf{Q8. D:} \% instances with a positive flag for a first-, \mbox{second-,} or third-person pronoun.}
    \label{fig:c-pronoun-dist}
\end{figure}

\begin{figure}[t!]
    \centering
    \includegraphics[width=\linewidth]{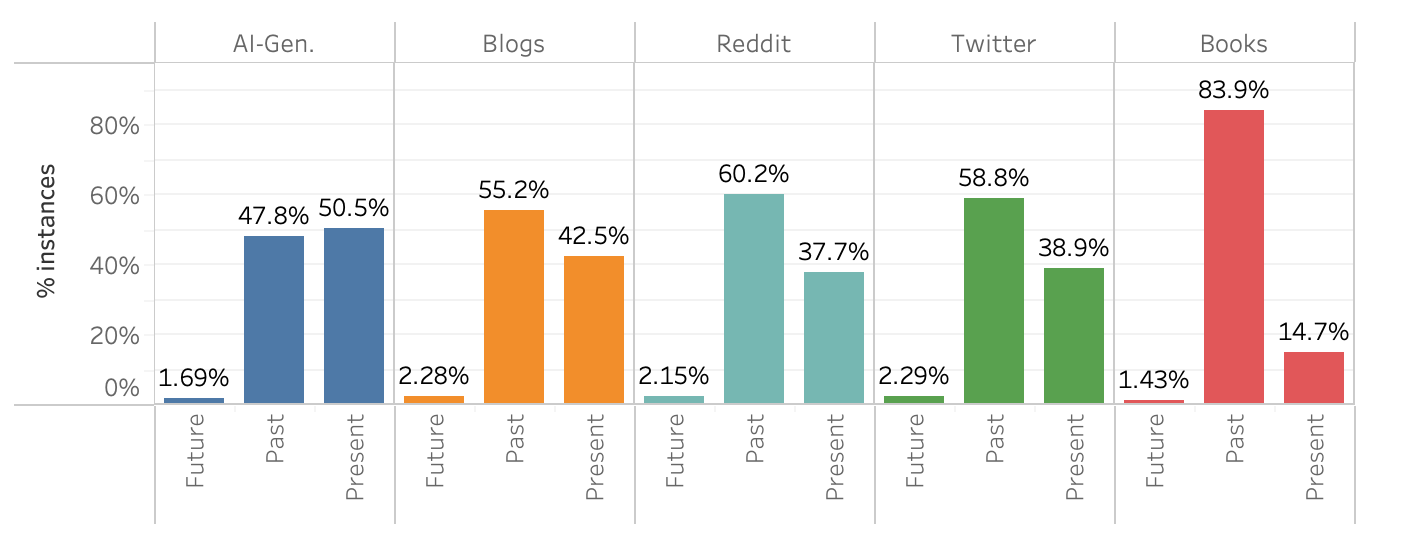}
    \caption{\textbf{Q8 Cognition:} \% instances with a positive flag for a specific tense verb or modal.}
    \label{fig:c-tense-dist}
\end{figure}

\noindent\textbf{Q8. D-Tense-Pronouns:} \textit{How do tense and pronoun usage differ by dataset?}\\
\textit{Motivation.} Tense and person reference capture narrative stance and interactional focus, which are core constructs in discourse and stylistics. For CSS, they reveal how people recount vs.\@ instruct, self-narrate vs.\@ address others. For NLP, they support interpretable style transfer and alignment of narrative voice in generation.\\
\textit{Results.} \Cref{fig:c-tense-dist} shows that future tense is rare (1.4--2.3\%) across all media. Past tense dominates Books (83.9\%) and remains high on Reddit (60.2\%), Twitter (58.8\%), and Blogs (55.2\%). AI-generated text slightly favors present (50.5\%) over past (47.8\%). Pronoun distributions (\Cref{fig:c-pronoun-dist}) indicate that first-person usage is most common on Reddit and Blogs (self-narration), whereas second-person occurs more on Twitter and in AI dialogues (addressivity).\\

\begin{figure}[t!]
    \centering
    \includegraphics[width=\linewidth]{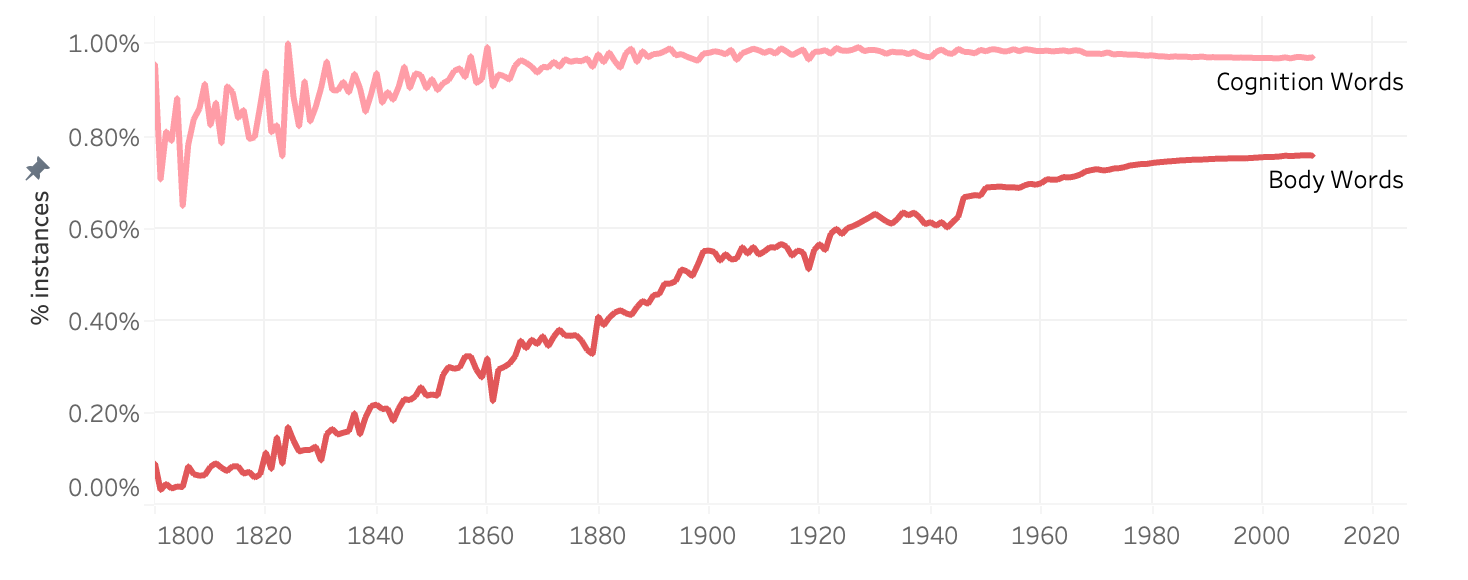}
    \caption{\textbf{Q9. B,C:} \% instances with body and cognition terms over 200 years in fiction books.}
    \label{fig:bc-books}
\end{figure}

\noindent\textbf{Q9. ABCDE:} 
\textit{How can \datasetName\ support cross-feature analyses across \textcolor{colorA}{\textbf{A}}ffect, \textcolor{colorB}{\textbf{B}}ody, \textcolor{colorC}{\textbf{C}}ognition, \textcolor{colorD}{\textbf{D}}emographics, and \textcolor{colorE}{\textbf{E}}motion?}\\
\textit{Motivation.} Many questions in affective and social science require connecting multiple linguistic signals at once; for example, how much embodied words are being used compared to cognitive words, and how that has changed over time. Cross‑feature views are a primary use case for \datasetName, which provides ready access to many features on the same instances and across instances over time.\\
\textit{Results.} As a demonstration, let us say we are interested in the rate at which body and cognition terms have been used over the last 200 years in English‑fiction books (Google Books 5‑grams, 1800–-2010). 
\Cref{fig:bc-books} plots the results.
Observe that the share of instances with at least one \emph{body} term rises more than an order of magnitude --- from about \textbf{0.05\%} circa 1800 to roughly \textbf{0.50\%} by \textbf{1900}. It then still continues to rise, albeit at a slower rate of increase,  to \textbf{0.62\%} in the \textbf{1930s--1940s}. %
There is a dip in the years of World War 2 (1938 to 1945), but it recovers back up to about \textbf{0.66\%} by \textbf{1950}. 
Mentions of body parts  plateaus to around \textbf{0.73--0.75\%} after \textbf{1980}. In contrast, the share of \emph{cognition} terms has always been higher than that of body terms and perhaps more importantly, it has remained fairly steady over time 
(hovering around \textbf{0.95--0.99\%} from the late 19th to mid‑20th century, with a slight softening to about \textbf{0.97--0.98\%} in the 1990s and 2000s). Consequently, the body‑to‑cognition ratio increase is stark from roughly \textbf{5\%} in \textbf{1800} to about \textbf{77\%} by the \textbf{2000s} (numbers not shown here) %
suggesting a long‑term rise in embodied language. 
Researchers can exploit various combinations of features in ABCDE to similarly explore research questions of interest.

\section{Conclusion}
In this work, we presented \datasetName, a collection of text datasets and linguistic annotations that capture features of interest at the intersection of language, affective science, cognition, and social science. We make all artifacts of \datasetName available, along with the associated code. %
Notably, several recent works already use \datasetName, including sentence-level annotations of social perception dimensions \cite{ayesh-etal-2026-warmth-competence-sentences}, analyses of emotion expression across the lifespan in social media \cite{teodorescu-etal-2026-emotion}, and analyses for large-scale lexical norms for warmth and trust \cite{mohammad-2025-words}, and anxiety \cite{mohammad-2026-worrylex}.

Additionally, we presented several representative analyses that reflect the potential of this dataset to answer fundamental research questions on the nature of affect, cognition, and behavior, using the rich signals encoded in language as data. Affective Scientists, for example, can use this resource to map the evolution of emotion word usage with age and social culture, study which emotions are most commonly associated with a focus on the self vs the other, the strength of the connection between bodily awareness and affective states, and the downstream impacts of such patterns on the mental and physical health of populations. The social media datasets in \datasetName can be used to answer central questions in computational social science on how social concepts are perceived and talked about in different sub-networks, and the change in these perceptions over time and in relation to real-world events. The inclusion of AI-generated texts enables a comparison of the ways in which large language models imitate or differ from human usage of natural language at an aggregate level, along dimensions such as emotion and cogitation, as well more fine-grained analyses of the dynamics of human-LLM conversations. 

Future work will add information for more affective, social perception, and moral dimensions: notably the care--harm, authority--subversion, loyalty--betrayal, fairness--unfairness, and purity--degradation dimensions from the NRC Moral Foundations Lexicon (based on the Moral Foundation Theory of \citet{haidt2007morality}).
We also hope that future work led by native speakers of various languages will include datasets in various languages.

\section*{Limitations and Ethical Considerations}
The scope and focus  of our work was on English corpora. Thus, much of the research that the corpora and features enable are relevant to English and North America. We discuss some of the relevant limitations and ethical considerations below. We hope this work will pave the way for creating similar resources in various other languages, led by speakers of those languages.
\begin{enumerate}
    \item English is the only language represented in our dataset. Many of the features of language use that we enumerate and operationalize here vary with language, and are reflective of underlying differences in cognition, affect, and behavior across population groups.
    \item A large proportion of the data in the datasets that we consolidate originate from North American users, and therefore is not representative of English-speaking populations from other countries, regions, and cultures. 
    \item The lexicons we use to quantify various features are also reflective of the usage patterns and biases of the annotators involved in the process, who are largely from the English-speaking populations of US and Europe.
    \item The broad use of static lexicons in our resource also may not account for the variations seen in language use with time, or across different domains of data. 
    \item Our feature set does not model syntax, discourse structure, or compositional effects. Even so, prior lexicon-based studies show that large-scale comparative trends can remain informative at aggregate scale \cite{teodorescu2023evaluating,ohman2024emotionarcs}.
\end{enumerate}
We refer readers to \citet{Mohammad23ethicslex} for an overview of the recommended practices in the usage of emotion lexicons, and \citet{mohammad-2022-ethics-sheet} for the ethical considerations that are relevant in the use of automatic emotion recognition tools.

We also enumerate below some of the many ethical considerations that apply to research in computational affective science and social science:
\begin{enumerate}
    \item Conclusions about language use, mental health, emotionality, etc. should be considered aggregate, population-level indicators of trends, rather than as tools for estimating or predicting the behavior of individuals.
    \item Conclusions drawn from our resource should be validated by other sources of data and measurement methods, such as controlled user studies.
  \item The creation of language datasets has several socio-cultural biases encoded in the process. What data, and whose data, is recorded, digitized, preserved, and published, either on the internet and in historical archives, is determined by social systems of power and influence. Language datasets, however large, should be critically interrogated to understand whose worldview is being represented, and to avoid over-claiming the generalizability of conclusions.
  \item The released instance-level annotations include demographic attributes such as age, gender, religion, occupation, and location. We strictly prohibit any commercial use of our dataset, or its use for profiling, targeting, or individual-level inference.
\end{enumerate}

\section*{Acknowledgments}
This work was supported by the Lower Saxony Ministry of Science and Culture and the VW Foundation. Thanks to Terry Ruas and Lars Kaesberg for early feedback on this work.

\section{Bibliographical References}
\bibliographystyle{lrec2026-natbib}
\bibliography{custom}

\clearpage
\onecolumn
\appendix
\section{Appendix: Feature Resources and Descriptive Statistics}
The following appendix lists details on the released features in \datasetName, including source resources and descriptive statistics.

\subsection{Feature Summary}
\label{app:feature-summary}
Table~\ref{tab:appendix-feature-summary} summarizes the released feature families, the label patterns that appear in the data files, the underlying resources, and the per-instance output types.
\footnotesize
\setlength{\tabcolsep}{4pt}
\renewcommand{\arraystretch}{1.12}
\begin{longtable}{@{}p{0.13\linewidth}p{0.33\linewidth}p{0.36\linewidth}p{0.12\linewidth}@{}}
\toprule
Family & Labels in release & Source resources and scope & Outputs\\
\midrule
\endfirsthead
\toprule
Family & Labels in release & Source resources and scope & Outputs\\
\midrule
\endhead
\bottomrule
\endfoot
\bottomrule
\caption{Released annotation features in \datasetName}\label{tab:appendix-feature-summary}
\endlastfoot
VAD & \texttt{NRCAvgValence}, \texttt{NRCAvgArousal}, \texttt{NRCAvgDominance}\newline \texttt{NRCHasHigh/LowValenceWord}\newline \texttt{NRCHasHigh/LowArousalWord}\newline \texttt{NRCHasHigh/LowDominanceWord}\newline \texttt{NRCCountHigh/LowValenceWords}\newline \texttt{NRCCountHigh/LowArousalWords}\newline \texttt{NRCCountHigh/LowDominanceWords} & NRC VAD \cite{vad-acl2018}; 19,971 entries with scalar Valence, Arousal, and Dominance scores & average, count, binary \\
Discrete Emotion & \texttt{NRCHas[Emotion]Word}\newline \texttt{NRCCount[Emotion]Words}\newline for anger, anticipation, disgust, fear, joy, sadness, surprise, trust, positive, and negative & NRC Emotion Lexicon \cite{Mohammad13}; 14,154 words with binary emotion and sentiment associations & count, binary \\
Anxiety / Calmness & \texttt{NRCHasAnxietyWord}, \texttt{NRCHasCalmnessWord}\newline \texttt{NRCAvgAnxiety}, \texttt{NRCAvgCalmness}\newline \texttt{NRCHasHighAnxietyWord}\newline \texttt{NRCCountHighAnxietyWords}\newline \texttt{NRCHasHighCalmnessWord}\newline \texttt{NRCCountHighCalmnessWords} & NRC WorryWords \cite{worrywords-emnlp2024}; 44,447 entries with signed anxiety or calmness scores & average, count, binary \\
Warmth / Trust & \texttt{NRCAvgMoralTrustWord}\newline \texttt{NRCAvgSocialWarmthWord}\newline \texttt{NRCAvgWarmthWord}\newline \texttt{NRCHasHigh/LowMoralTrustWord}\newline \texttt{NRCHasHigh/LowSocialWarmthWord}\newline \texttt{NRCHasHigh/LowWarmthWord}\newline \texttt{NRCCountHigh/LowMoralTrustWord}\newline \texttt{NRCCountHigh/LowSocialWarmthWord}\newline \texttt{NRCCountHigh/LowWarmthWord} & NRC MoralTrust, SocialWarmth, and CombinedWarmth \cite{mohammad-2025-words}; 31,456--31,573 entries across the three lexicons & average, count, binary \\
BPMs & \texttt{HasBPM}, \texttt{MyBPM}, \texttt{YourBPM}\newline \texttt{HerBPM}, \texttt{HisBPM}, \texttt{TheirBPM} & Released BPM inventory; 292 entries in file order, with duplicate \textit{toe} retained; compiled from public body-part lists and linked body-language work \cite{zhuang-etal-2024-heart,wu2025language} & binary, list \\
Cognition & one \texttt{COGHas...Word} binary per category, including\newline \texttt{COGHasAnalyzingEvaluatingWord}\newline \texttt{COGHasCreativityIdeationWord}\newline \texttt{COGHasGeneralCognitionWord}\newline \texttt{COGHasLearningUnderstandingWord}\newline \texttt{COGHasExplanationArticulationWord} & Released cognition inventory; 12 categories and 98 terms & binary \\
Demographics & \texttt{Author}, \texttt{DMGMajorityBirthyear}\newline \texttt{DMGRawBirthyearExtractions}\newline \texttt{DMGRawExtractedAge}, \texttt{DMGAgeAtPost}\newline \texttt{DMGRawExtractedGender}\newline \texttt{DMGRawExtractedCity}\newline \texttt{DMGCountryMappedFrom}\newline \texttt{ExtractedCity}\newline \texttt{DMGRawExtractedCountry}\newline \texttt{DMGRawExtractedReligion}\newline \texttt{DMGMainReligionMappedFrom}\newline \texttt{ExtractedReligion}\newline \texttt{DMGMainCategoryMappedFrom}\newline \texttt{ExtractedReligion}\newline \texttt{DMGRawExtractedOccupation}\newline \texttt{DMGSOCTitleMappedFrom}\newline \texttt{ExtractedOccupation} & Regex extraction plus controlled vocabularies: 6 age regexes, SOC occupations, 69 gender terms, 231 countries, top 50k GeoNames cities, and 157 religion rows & extracted strings, mapped labels, numeric age \\
Pronouns & binary \texttt{PRNHas...} binarys from\newline \texttt{PRNHasI} to \texttt{PRNHasTheirs} across first-, second-, and third-person forms & Released pronoun inventory covering first-person singular and plural, second person, and third-person feminine, masculine, and plural or neutral forms & binary \\
Time / Tense & \texttt{TIMEHasPastVerb}, \texttt{TIMECountPastVerbs}\newline \texttt{TIMEHasPresentVerb}\newline \texttt{TIMECountPresentVerbs}\newline \texttt{TIMEHasFutureModal}\newline \texttt{TIMECountFutureModals}\newline \texttt{TIMEHasPresentNoFuture}\newline \texttt{TIMEHasFutureReference} & UniMorph English plus future-modal rules; tense features derived from present, past, and future-oriented forms & count, binary \\
General & \texttt{WordCount} & tokenized text stream for each instance & count \\
\end{longtable}

\subsection{Dataset Summary Statistics}
\label{app:dataset-stats}
Table~\ref{tab:appendix-dataset-summary} reports the instance definition and lexical statistics for each source. MTLD (Measure of Textual Lexical Diversity) estimates the average token span over which lexical variation remains above a fixed threshold \cite{mccarthy2010mtld}. MATTR (Moving-Average Type-Token Ratio) averages the type-token ratio over a sliding window to reduce text-length sensitivity \cite{covington2010mattr}. For both metrics, higher values indicate greater lexical diversity. MATTR ranges from 0 to 1, while MTLD has no fixed upper bound. A few contrasts are notable. Twitter has the shortest instances on average (20.83 tokens) because of its limited posting length, but the highest lexical-diversity values (MTLD 252.31, MATTR 0.88), whereas Books are fixed-length 5-grams and therefore have the lowest diversity values (MTLD 6.23, MATTR 0.09). Blogs and AI-generated texts are much longer on average (199.57 and 258.65 tokens), but AI-generated text remains substantially less lexically diverse than either Blogs or Reddit.
\begin{table*}[h!]
\small
\centering
\setlength{\tabcolsep}{5pt}
\renewcommand{\arraystretch}{1.05}
\begin{tabularx}{\linewidth}{@{}>{\raggedright\arraybackslash}p{0.13\linewidth}>{\raggedleft\arraybackslash}p{0.13\linewidth}>{\raggedright\arraybackslash}X>{\raggedleft\arraybackslash}p{0.11\linewidth}>{\raggedleft\arraybackslash}p{0.11\linewidth}>{\raggedleft\arraybackslash}p{0.10\linewidth}>{\raggedleft\arraybackslash}p{0.10\linewidth}@{}}
\toprule
Source & Instances & Instance definition & \shortstack{Mean\\tokens} & \shortstack{Median\\tokens} & MTLD & MATTR \\
\midrule
Twitter & 45.2M & one post & 20.83 & 16.00 & 252.31 & 0.88 \\
Reddit & 78.5M & one post & 117.78 & 72.00 & 91.52 & 0.81 \\
Books & 177.1M (1.74M unique) & one 5-gram occurence & 5.00 & 5.00 & 6.23 & 0.09 \\
Blogs & 34.2M & one sentence & 199.57 & 96.00 & 107.38 & 0.83 \\
AI-generated & 68.9M & one AI response & 258.65 & 143.00 & 30.60 & 0.67 \\
\bottomrule
\end{tabularx}
\caption{Per-source descriptive statistics. MTLD and MATTR are computed over the concatenated token stream for each source (MATTR window = 50; MTLD threshold = 0.72). Higher values on both metrics indicate greater lexical diversity; MATTR ranges from 0 to 1, while MTLD has no fixed upper bound.}
\label{tab:appendix-dataset-summary}
\end{table*}

Table~\ref{tab:appendix-demographic-coverage} reports demographic coverage as counts in thousands and percentages of total source instances with age, occupation, gender, country, city, and religion annotations.
\begin{table*}[h!]
\small
\centering
\setlength{\tabcolsep}{5pt}
\renewcommand{\arraystretch}{1.05}
\begin{tabularx}{\linewidth}{@{}>{\raggedright\arraybackslash}m{0.16\linewidth}*{6}{>{\raggedleft\arraybackslash}X}@{}}
\toprule
Source & Age & Occupation & Gender & Country & City & Religion \\
\midrule
Twitter & \makecell[r]{2,150k\\{\scriptsize (4.8\%)}} & \makecell[r]{525k\\{\scriptsize (1.2\%)}} & \makecell[r]{502k\\{\scriptsize (1.1\%)}} & \makecell[r]{4,900k\\{\scriptsize (10.8\%)}} & \makecell[r]{3,496k\\{\scriptsize (7.7\%)}} & \makecell[r]{334k\\{\scriptsize (0.7\%)}} \\
Reddit & \makecell[r]{33,495k\\{\scriptsize (42.7\%)}} & \makecell[r]{3,041k\\{\scriptsize (3.9\%)}} & \makecell[r]{3,451k\\{\scriptsize (4.4\%)}} & \makecell[r]{12,558k\\{\scriptsize (16.0\%)}} & \makecell[r]{8,653k\\{\scriptsize (11.0\%)}} & \makecell[r]{2,283k\\{\scriptsize (2.9\%)}} \\
\bottomrule
\end{tabularx}
\caption{Demographic coverage with self-disclosure annotations. Cells show the number of instances in thousands, together with the percentage of total source instances for that source.}
\label{tab:appendix-demographic-coverage}
\end{table*}

\subsection{Full BPM Inventory}
\label{app:bpm}
Table~\ref{tab:appendix-bpm} shows the BPM words used in alphabetical order.
\footnotesize
\begin{table}[h!]
\centering
\begin{tabularx}{\textwidth}{@{}p{0.18\textwidth}X@{}}
\toprule
First-Letter Group & Terms \\
\midrule
A-C & adam's apple, abdomen, abdominal, adenoid, adenoids, adrenal gland, alvine, anal, anatomy, ankle, anus, appendicular, appendix, arch, arm, armpit, arterial, artery, axilla, axillary, back, ball of the foot, belly, belly button, toe, bladder, blood, blood vessels, body, bone, brachial, brachium, brain, breast, buttocks, caecum, calf, capillary, capital, caput, cardiac, carpal, carpus, cartilage, cell, cerebral, cervical, cervical vertebrae, cervix, cheek, chest, chin, ciliary, cilium, circulatory system, clavicle, clitoral, clitoris, coccyx, collar bone, colon, colonic, crural, crus, cubital, cutaneous, cutis \\
D-H & diaphragm, digestive system, digital, dorsal, duodenal, duodenum, ear, ear lobe, elbow, encephalon, endocrine system, epiglottis, erythrocyte, esophagus, eye, eyebrow, eyelash, eyelashes, eyelid, face, fallopian tubes, feet, femur, fibula, filling, finger, fingernail, fist, follicle, foot, forearm, forehead, foreskin, frontal, gallbladder, gastric, gena, genal, genicular, genu, gingiva, gingival, gland, glands, glottal, glottic, glottis, groin, gullet, gum, gums, hair, half-moon, hamstring, hand, head, heart, heel, hepatic, hip, humerus \\
I-M & ileum, immune system, index finger, inguinal, instep, intestine, intestines, iris, jaw, jejunum, kidney, knee, knuckle, labial, labyrinth, laryngeal, larynx, leg, leucocyte, ligament, lingua, lip, liver, lobe, loin, lumbar, lumbar vertebrae, lumbus, lung, lungs, lymph node, lymphocyte, mandible, manual, manus, metacarpal, metatarsal, midriff, molar, mouth, muscle \\
N-R & nail, nape, naris, nasal, nates, navel, neck, nerve, nerves, neural, nipple, nose, nostril, nucha, occipital, occiput, oesophagus, organs, ovarian, ovary, oviduct, palm, palpebral, pancreas, pancreatic, patella, pectoral, pedal, pelvis, penile, penis, phalanges, pharyngeal, pharynx, pinky, pituitary, plantar, pollex, popliteal, pore, prepuce, pubes, pubic, pulmonary, pupil, radius, rectal, rectum, red blood cells, respiratory system, ribcage, ribs \\
S-W & sacrum, scalp, scapula, scrotum, senses, shin, shoulder, shoulder blade, skeleton, skin, skull, sole, spinal column, spinal cord, spine, spleen, sternum, stomach, stomatic, superciliary, sural, talus, tarsal, teeth, temple, temporal, tendon, testes, testicle, testicular, thigh, thoracic, thorax, throat, thumb, thyroid, tibia, tissue, toe, toenail, tongue, tonsil, tonsils, tooth, torso, trachea, trunk, ulna, umbilical, umbilicus, ureter, urethra, urinary system, uterine, uterus, uvula, vagina, vaginal, vein, vena, venous, venter, ventral, vertebra, vesical, vulva, waist, white blood cells, windpipe, womb, wrist \\
\bottomrule
\end{tabularx}
\caption{Released BPM inventory in file order.}
\label{tab:appendix-bpm}
\end{table}
\clearpage

\subsection{Cognition Word Inventory}
\label{app:cog}
Table~\ref{tab:appendix-cog} shows the cognition word inventory.
\footnotesize
\begin{table}[h!]
\centering
\begin{tabularx}{\textwidth}{@{}p{0.3\textwidth}X@{}}
\toprule
Category & Terms \\
\midrule
Analyzing \& Evaluating & analyze, appraise, assess, critique, diagnose, differentiate, discern, discriminate, distinguish, evaluate, investigate, self-evaluate, test \\
Creativity \& Ideation & brainstorm, conceptualize, create, diverge, evolve, fantasize, ideate, imagine, innovate, invent, pretend, visualize \\
General Cognition & contemplate, deliberate, focus, introspect, reason, reflect, regulate, ruminate \\
Learning \& Understanding & accommodate, assimilate, associate, comprehend, empathize, explore, grasp, internalize, learn, study, understand \\
Decision Making \& Judging & calculate, choose, decide, deduce, determine, estimate, infer, judge, prioritize, resolve \\
Problem Solving & plan, revise, solve, strategize, troubleshoot \\
Higher-Order Thinking & abstract, categorize, classify, generalize, hypothesize, interpret, synthesize \\
Confused or Uncertain Thinking & doubt, self-question \\
Memory \& Recall & consolidate, encode, forget, memorize, recall, rehearse, remember, retrieve \\
Perception \& Observation & detect, identify, label, notice, observe, perceive, recognize, scan, spot, trace \\
Prediction \& Forecasting & anticipate, forecast, forethink, predict, project \\
Explanation \& Articulation & articulate, define, describe, discuss, elaborate, explain, verbalize \\
\bottomrule
\end{tabularx}
\caption{Cognition word inventory grouped by category.}
\label{tab:appendix-cog}
\end{table}
\clearpage

\subsection{Demographic Regex Overview}
\label{app:dmg-regex}
Table~\ref{tab:appendix-dmg-regex} summarizes the self-disclosure regex patterns used by the demographic extractor.
\begin{table}[h!]
\centering
\begin{tabular}{lp{10cm}}
\toprule
\textbf{Attribute} & \textbf{Pattern(s)} \\
\midrule
age & (i) ``I am/I'm \textbf{XX} years old'' (explicit age statement) \\
& (ii) ``I am/I'm \textbf{XX}'' (+ end of string, punct./conj.) \\
& (iii) ``I was/am born in \textbf{YYYY}'' (four-digit year) \\
& (iv) ``I was/am born in '\textbf{YY}'' (two-digit year with apostrophe) \\
& (v) ``I was born on DD Month \textbf{YYYY}'' (full date format) \\
& (vi) ``I was born on \textbf{MM/DD/YYYY}'' (and similar date formats) \\
\hline
occupation & (i) ``I am/I'm (a/an/the) \textbf{[Occupation]} (at \textbf{[Company]}) (+ punt./conj.)'' \\
& (ii) ``I work as (a/an/the) \textbf{[Occupation]} (at \textbf{[Company]})'' \\
& (iii) ``My job/occupation/role is (to be) (a/an/as) \textbf{[Occupation]}'' \\
& (iv) ``I'm (currently) employed as (a/an/the) \textbf{[Occupation]}'' \\
\hline
gender & (i) ``I am/I'm (a/an) \textbf{[Gender]} (+ punctuation/conjunction)'' \\
& (ii) ``I identify as (a/an) \textbf{[Gender]}'' \\
& (iii) ``My gender (identity) is \textbf{[Gender]}'' \\
& (iv) ``I'm/I am (a) (transgender/trans) \textbf{[Gender]}'' \\
\hline
country & (i) ``I am/I'm from (the) \textbf{[Country]}'' (current primary location) \\
& (ii) ``I am/I'm (a/an/the) \textbf{[Nationality]} (+ punt./conj.)'' \\
& (iii) ``I live in/at (the) \textbf{[Country]}'' \\
& (iv) ``I come from (the) \textbf{[Country]}'' \\
& (v) ``My nationality/citizenship is \textbf{[Nationality]}'' \\
& (vi) ``I was/am born and raised in (the) \textbf{[Country]}'' \\
& (vii) ``I am/I'm originally from (the) \textbf{[Country]}'' \\
\hline
city & (i) ``I am/I'm from \textbf{[City]} (+ punctuation/conjunction)'' \\
& (ii) ``I live in/at \textbf{[City]}'' \\
& (iii) ``I'm/I am (currently) residing/based in \textbf{[City]}'' \\
& (iv) ``My (current/home) city/town is \textbf{[City]}'' \\
& (v) ``I (grew up / was raised) in \textbf{[City]}'' \\
\hline
religion & (i) ``I am/I'm (a/an/a practicing) \textbf{[Religion]} (+ context)'' \\
& (ii) ``My religion/faith is \textbf{[Religion]}'' \\
& (iii) ``I (actively) practice \textbf{[Religion]}'' \\
& (iv) ``I am/I'm a follower of \textbf{[Religion]}'' \\
& (v) ``I converted to \textbf{[Religion]}'' \\
& (vi) ``I was raised/born (as a/an) \textbf{[Religion]}'' \\
& (vii) ``I identify as \textbf{[Religion]} / identify with \textbf{[Religion]}'' \\
\bottomrule
\end{tabular}
\caption{Extraction patterns for labeling social media users with demographic attributes. For a full list of \textbf{[Occupation]}, \textbf{[Genders]}, \textbf{[Country]}, \textbf{[City]}, and \textbf{[Religion]}, please refer to the GitHub repository.}
\label{tab:appendix-dmg-regex}
\end{table}

\clearpage

\makeAIUsageCard

\clearpage
\onecolumn
\hypertarget{annotation}{}
\pagestyle{empty}
\lstset{
  basicstyle=\footnotesize\ttfamily,
  breaklines=true,
  breakatwhitespace=false,
  columns=flexible,
  numbers=none
}

\definecolor{Primary}{RGB}{59, 130, 246}    %
\definecolor{PrimaryDark}{RGB}{30, 64, 175} %
\definecolor{LightBg}{RGB}{239, 246, 255}   %
\definecolor{TextDark}{RGB}{31, 41, 55}     %
\definecolor{TextMuted}{RGB}{107, 114, 128} %

\begin{tikzpicture}[remember picture, overlay]
  \fill[Primary] ([xshift=0cm,yshift=0cm]current page.north west) rectangle ([xshift=\paperwidth,yshift=-0.4cm]current page.north west);
\end{tikzpicture}

\vspace{0.8cm}
\begin{center}
  {\fontsize{22}{26}\selectfont\sffamily\bfseries \textcolor{PrimaryDark}{CiteAssist}}\\[0.2em]
  {\Large\sffamily\scshape \textcolor{TextMuted}{Citation Sheet}}\\[0.8em]
  {\small\sffamily Generated with \href{https://citeassist.uni-goettingen.de/}{\textcolor{Primary}{\texttt{citeassist.uni-goettingen.de}}}
  \CiteAssistCite{}
  }\end{center}

\begin{center}
\vspace{1em}
\begin{tikzpicture}
\draw[Primary, line width=0.6pt] (0,0) -- (\textwidth,0);
\end{tikzpicture}
\vspace{1.2em}
\end{center}

\begin{tcolorbox}[enhanced,
                 frame hidden,
                 boxrule=0pt,
                 borderline west={2pt}{0pt}{Primary},
                 colback=LightBg,
                 sharp corners,
                 breakable,
                 fonttitle=\sffamily\bfseries\large,
                 coltitle=Primary,
                 title=BibTeX Entry,
                 attach title to upper={\vspace{0.2em}\par},
                 left=12pt]
\lstset{
    inputencoding = utf8,  %
    extendedchars = true,  %
    literate      =        %
      {á}{{\'a}}1  {é}{{\'e}}1  {í}{{\'i}}1 {ó}{{\'o}}1  {ú}{{\'u}}1
      {Á}{{\'A}}1  {É}{{\'E}}1  {Í}{{\'I}}1 {Ó}{{\'O}}1  {Ú}{{\'U}}1
      {à}{{\`a}}1  {è}{{\`e}}1  {ì}{{\`i}}1 {ò}{{\`o}}1  {ù}{{\`u}}1
      {À}{{\`A}}1  {È}{{\`E}}1  {Ì}{{\`I}}1 {Ò}{{\`O}}1  {Ù}{{\`U}}1
      {ä}{{\"a}}1  {ë}{{\"e}}1  {ï}{{\"i}}1 {ö}{{\"o}}1  {ü}{{\"u}}1
      {Ä}{{\"A}}1  {Ë}{{\"E}}1  {Ï}{{\"I}}1 {Ö}{{\"O}}1  {Ü}{{\"U}}1
      {â}{{\^a}}1  {ê}{{\^e}}1  {î}{{\^i}}1 {ô}{{\^o}}1  {û}{{\^u}}1
      {Â}{{\^A}}1  {Ê}{{\^E}}1  {Î}{{\^I}}1 {Ô}{{\^O}}1  {Û}{{\^U}}1
      {œ}{{\oe}}1  {Œ}{{\OE}}1  {æ}{{\ae}}1 {Æ}{{\AE}}1  {ß}{{\ss}}1
      {ẞ}{{\SS}}1  {ç}{{\c{c}}}1 {Ç}{{\c{C}}}1 {ø}{{\o}}1  {Ø}{{\O}}1
      {å}{{\aa}}1  {Å}{{\AA}}1  {ã}{{\~a}}1  {õ}{{\~o}}1 {Ã}{{\~A}}1
      {Õ}{{\~O}}1  {ñ}{{\~n}}1  {Ñ}{{\~N}}1  {¿}{{?\`}}1  {¡}{{!\`}}1
      {„}{\quotedblbase}1 {“}{\textquotedblleft}1 {–}{$-$}1
      {°}{{\textdegree}}1 {º}{{\textordmasculine}}1 {ª}{{\textordfeminine}}1
      {£}{{\pounds}}1  {©}{{\copyright}}1  {®}{{\textregistered}}1
      {«}{{\guillemotleft}}1  {»}{{\guillemotright}}1  {Ð}{{\DH}}1  {ð}{{\dh}}1
      {Ý}{{\'Y}}1    {ý}{{\'y}}1    {Þ}{{\TH}}1    {þ}{{\th}}1    {Ă}{{\u{A}}}1
      {ă}{{\u{a}}}1  {Ą}{{\k{A}}}1  {ą}{{\k{a}}}1  {Ć}{{\'C}}1    {ć}{{\'c}}1
      {Č}{{\v{C}}}1  {č}{{\v{c}}}1  {Ď}{{\v{D}}}1  {ď}{{\v{d}}}1  {Đ}{{\DJ}}1
      {đ}{{\dj}}1    {Ė}{{\.{E}}}1  {ė}{{\.{e}}}1  {Ę}{{\k{E}}}1  {ę}{{\k{e}}}1
      {Ě}{{\v{E}}}1  {ě}{{\v{e}}}1  {Ğ}{{\u{G}}}1  {ğ}{{\u{g}}}1  {Ĩ}{{\~I}}1
      {ĩ}{{\~\i}}1   {Į}{{\k{I}}}1  {į}{{\k{i}}}1  {İ}{{\.{I}}}1  {ı}{{\i}}1
      {Ĺ}{{\'L}}1    {ĺ}{{\'l}}1    {Ľ}{{\v{L}}}1  {ľ}{{\v{l}}}1  {Ł}{{\L{}}}1
      {ł}{{\l{}}}1   {Ń}{{\'N}}1    {ń}{{\'n}}1    {Ň}{{\v{N}}}1  {ň}{{\v{n}}}1
      {Ő}{{\H{O}}}1  {ő}{{\H{o}}}1  {Ŕ}{{\'{R}}}1  {ŕ}{{\'{r}}}1  {Ř}{{\v{R}}}1
      {ř}{{\v{r}}}1  {Ś}{{\'S}}1    {ś}{{\'s}}1    {Ş}{{\c{S}}}1  {ş}{{\c{s}}}1
      {Š}{{\v{S}}}1  {š}{{\v{s}}}1  {Ť}{{\v{T}}}1  {ť}{{\v{t}}}1  {Ũ}{{\~U}}1
      {ũ}{{\~u}}1    {Ū}{{\={U}}}1  {ū}{{\={u}}}1  {Ů}{{\r{U}}}1  {ů}{{\r{u}}}1
      {Ű}{{\H{U}}}1  {ű}{{\H{u}}}1  {Ų}{{\k{U}}}1  {ų}{{\k{u}}}1  {Ź}{{\'Z}}1
      {ź}{{\'z}}1    {Ż}{{\.Z}}1    {ż}{{\.z}}1    {Ž}{{\v{Z}}}1  {ž}{{\v{z}}}1
  }
\begin{lstlisting}
@inproceedings{wahle-etal-2026-abcde,
  address={Palma de Mallorca, Spain},
  author={Wahle, Jan Philip and Vishnubhotla, Krishnapriya, Gipp, Bela and Mohammad, Saif M.},
  booktitle={Proceedings of the 1st Workshop on Computational Affective Science (CAS 2026)},
  month={may},
  publisher={European Language Resources Association (ELRA)},
  title={Affect, Body, Cognition, Demographics, and Emotion: The ABCDE of Text Features for Computational Affective Science},
  year={2026}
}
\end{lstlisting}
\end{tcolorbox}

\vfill
\begin{tikzpicture}
\draw[Primary!40, line width=0.4pt] (0,0) -- (\textwidth,0);
\end{tikzpicture}
\begin{center}
\small\sffamily\textcolor{TextMuted}{Generated \today}
\end{center}

\end{document}